\documentclass[runningheads]{llncs}
\usepackage{graphicx}

\usepackage{tikz}
\usepackage{comment}
\usepackage{amsmath,amssymb} 
\usepackage{color}

\usepackage{graphicx}
\usepackage{amsmath}
\usepackage{amssymb}
\usepackage{booktabs}
\usepackage{xspace}
\usepackage{interval}
\usepackage{bm}
\usepackage{mathtools}
\usepackage{footmisc}

\usepackage{siunitx}
\usepackage{bbold}
\usepackage{multirow}
\usepackage{bbm}
\usepackage{amssymb}
\usepackage{bbding}
\usepackage{cite}
\usepackage{xspace}

\makeatletter
\DeclareRobustCommand\onedot{\futurelet\@let@token\@onedot}
\def\@onedot{\ifx\@let@token.\else.\null\fi\xspace}

\def\eg{\emph{e.g}\onedot} 
\def\ie{\emph{i.e}\onedot}

\def\etal{\emph{et al}\onedot}
\makeatother

\newcommand{\fakeparagraph}[1]{\textbf{#1}}

\definecolor{citecolor}{RGB}{0, 113, 188}

\newcommand*{\system}{\texttt{Semiformer}\@\xspace}
\newcommand*{\vanilla}{\texttt{Vanilla}\@\xspace}
\newcommand*{\convl}{\texttt{Conv-labeled}\@\xspace}

\newcommand*{\supervised}{\texttt{Sup.}\@\xspace}

\makeatletter
\newcommand\figcaption{\def\@captype{figure}\caption}
\newcommand\tabcaption{\def\@captype{table}\caption}
\makeatother

\definecolor{green_im}{rgb}{0.1, 0.55, 0.3}
\definecolor{citecolor}{RGB}{0, 113, 188}
\usepackage[pagebackref=true,breaklinks=true,letterpaper=true,colorlinks,bookmarks=false, citecolor=citecolor]{hyperref}

\usepackage{orcidlink}
\usepackage[capitalize]{cleveref}
\crefname{section}{Sec.}{Secs.}
\Crefname{section}{Section}{Sections}
\Crefname{table}{Table}{Tables}
\crefname{table}{Tab.}{Tabs.}

\newcommand{\inet}{{\scshape ImageNet}\xspace}
\newcommand{\places}{{\scshape Places205}\xspace}

\usepackage[accsupp]{axessibility}  

\usepackage[normalem]{ulem}

\newcommand{\Rise}[1]{\textcolor{green_im}{\small{(\bf $\uparrow$#1})}\xspace}

\begin{document}
\pagestyle{headings}
\mainmatter
\def\ECCVSubNumber{3393}  
\sloppy
\newcommand\blfootnote[1]{%
  \begingroup
  \renewcommand\thefootnote{}\footnote{#1}%
  \addtocounter{footnote}{-1}%
  \endgroup
}

\title{Semi-Supervised Vision Transformers} 



\titlerunning{Semi-Supervised Vision Transformers}
%
\author{Zejia Weng$^{1,2*}\orcidlink{0000-0001-9706-6484}$,\, Xitong Yang$^ {3*}$\orcidlink{0000-0003-4372-241X},\, Ang Li$^{4}$\orcidlink{0000-0003-4795-8374}, \\ Zuxuan Wu$^{1,2\dagger}$ \orcidlink{0000-0002-8689-5807},\, Yu-Gang Jiang$^{1,2\dagger}$\orcidlink{0000-0002-1907-8567}
}

\authorrunning{Zejia Weng et al.}
%
\institute{$^{1}$~Shanghai Key Lab of Intell. Info. Processing, School of CS, Fudan University \\
$^{2}$~Shanghai Collaborative Innovation Center on Intelligent Visual Computing\\
$^{3}$~Meta AI $^{4}$~Baidu Apollo\\
}

\maketitle

\begin{abstract}
\blfootnote{$^*$Equal contributions. $^{\dagger}$Corresponding author.}
We study the training of Vision Transformers for semi-supervised image classification. Transformers have recently demonstrated impressive performance on a multitude of supervised learning tasks. Surprisingly, we show Vision Transformers perform significantly worse than Convolutional Neural Networks when only a small set of labeled data is available. Inspired by this observation, we introduce a joint semi-supervised learning framework, \system, which contains a transformer stream, a convolutional stream and a carefully designed fusion module for knowledge sharing between these streams. The convolutional stream is trained on limited labeled data and further used to generate pseudo labels to supervise the training of the transformer stream on unlabeled data. Extensive experiments on ImageNet demonstrate that \system achieves 75.5\% top-1 accuracy, outperforming the state-of-the-art by a clear margin. In addition, we show, among other things, \system is a general framework that is compatible with most modern transformer and convolutional neural architectures. Code is available at \href{https://github.com/wengzejia1/Semiformer}{https://github.com/wengzejia1/Semiformer}.

\end{abstract}

\section{Introduction}

Vision transformers (ViT) have achieved remarkable performance recently on a variety of supervised computer vision tasks \cite{dosovitskiy2020image,heo2021rethinking,liu2021swin}. Their success is largely fueled by high capacity models with self-attention layers trained on massive data. However, it is not always feasible to collect sufficient annotated data in many real world applications. 
When only a small number of labeled samples are provided, semi-supervised learning (SSL) \cite{zhu2005semi,chapelle2009semi} is a powerful paradigm to achieve better performance by leveraging a huge amount of unlabeled data. Despite the success of Vision Transformers in fully supervised scenarios, the understanding of its effectiveness in SSL is still an empty space.

\begin{figure}[t!]
\centering
  \includegraphics[width=0.9\linewidth]{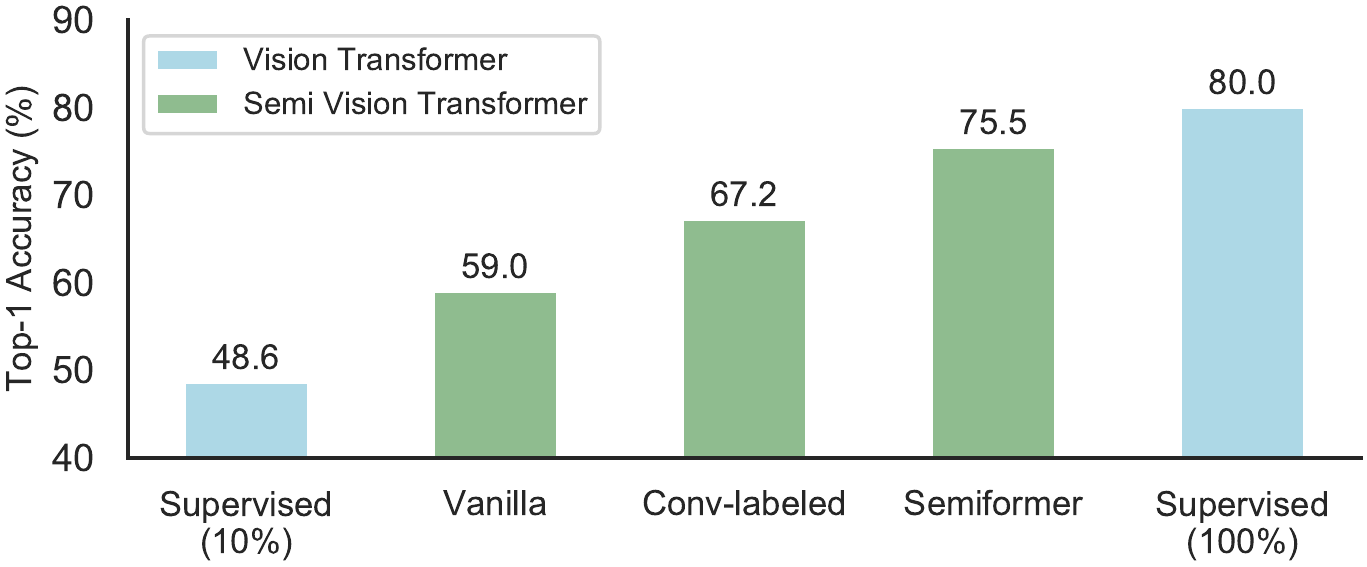}
  \caption{Three semi-supervised vision transformers using 10\% labeled and 90\% unlabeled data (colored in green) vs. fully supervised vision transformers (colored in blue) using 10\% and 100\% labeled data. Our  approach {\system} achieves competitive performance, 75.5\% top-1 accuracy.} 
\label{fig:CTcompare}
\end{figure}

We perform a series of studies with Vision Transformers (ViT)~\cite{dosovitskiy2020image} in the semi-supervised learning (SSL) setting on ImageNet. Surprisingly, the results show that simply training a ViT using a popular SSL approach, FixMatch~\cite{sohn2020fixmatch}, still leads to much worse performance than a CNN trained even without FixMatch. We believe this results from the fact that pseudo labels from CNNs are more accurate, possibly due to their encoded inductive bias.

To validate our hypothesis, we use CNNs to produce pseudo labels for the joint semi-supervised training of CNNs and transformers.
By doing so, we are able to significantly improve the top-1 accuracy of the ViT by 8+\% (c.f. \convl and \vanilla in \cref{fig:CTcompare}). This highlights that labels derived from CNNs are also helpful for training transformers under the SSL setting. While pseudo labels from CNNs are effective, the final ViT is still slightly weaker than the ``teacher" CNN.
We posit that simply performing pseudo labeling (PL) with CNNs to derive supervisory signals for transformers is not sufficient. Instead, we hypothesize that a joint knowledge sharing mechanism at the architecture level is required to fully explore knowledge in CNNs. 

In light of these, we introduce a novel semi-supervised learning framework for Vision Transformers, which we term as {\system}. In particular, \system composes of a convolutional stream and a transformer stream. It leverages labels produced by CNNs as supervisory signals to train the CNN and the transformers jointly using a popular SSL strategy. The two streams are further connected with a cross-stream feature interaction module, enabling streams to complement each other.  Benefited from more accurate labels and the interaction design, \system can be readily used for SSL. 

We conduct extensive experiments to evaluate \system.  In particular, \system achieves 75.5\% top-1 accuracy on ImageNet and outperforms the state-of-the-art using 10\% of labeled samples. We also show \system outperforms alternative methods by clear margins under different labeling ratios. In addition, we empirically demonstrate \system is a generic framework compatible with modern CNN and transformer architectures. We also provide qualitative evidence that \system is better than ViTs in the SSL setting. \newline

\textbf{Contributions. } Our contributions are three-folded:
\begin{enumerate}
\itemsep 3pt
    \item We are the first to investigate the application of Vision Transformers for semi-supervised learning. We reveal that Vision Transformers perform poorly when labeled samples are limited, yet they can be improved by utilizing unlabeled data together with the help from Convolutional neural networks. 
    \item We propose a generic framework \system for the semi-supervised learning of Vision Transformers, which not only explores predictions as supervisory signals but also feature-level clues from CNNs to improve the ViTs in the low-data learning regime.
    \item We perform extensive experiments and studies to evaluate \system.  \system achieves 75.5\% top-1 accuracy on ImageNet and outperforms state-of-the-art methods in semi-supervised learning. Additional ablation studies are further conducted to understand its effectiveness. 
\end{enumerate}

\section{Related work}

\textbf{Vision Transformers.} A variety of Vision Transformers~\cite{dosovitskiy2020image, yuan2021tokens, heo2021rethinking, wang2021pyramid, touvron2021training,wang2022bevt,wang2021efficient,li2022contextual} have refreshed the state-of-the-art performance on ImageNet, demonstrating their powerful representation capability in solving vision tasks. Among them, the Vision Transformer (ViT) \cite{dosovitskiy2020image} is the first to prove that purely using the transformer structure can perform well on image classification tasks. It divides each image into a sequence of patches and then applies multiple transformer layers~\cite{vaswani2017attention} to model their global relations.  T2T-ViT \cite{yuan2021tokens} recursively aggregates neighboring tokens into one token for better modeling of local structures such as edges and lines among neighboring pixels, which outperforms ResNets \cite{he2016deep} and also achieves comparable performance to light CNNs by directly training on ImageNet. Swin Transformer \cite{liu2021swin} creates a shifted windowing scheme cooperated with stacked local transformers for better information interaction among patches. With the continuous improvements of Vision Transformers, transformer based networks have achieved higher accuracy on medium-scale and large-scale datasets. Although transformers have been proven effective at solving visual tasks, it is known inferior to some CNNs when training from scratch on small-sized datasets mainly because ViTs lack image-specific inductive bias~\cite{dosovitskiy2020image}. 

Touvron \etal \cite{touvron2021training} distill the knowledge of CNNs to ViTs, easing the training process of transformers to be more data efficient. The hard distillation idea is similar to the pseudo label approach in SSL. However, it differs from our work in that the teacher model in distillation is pre-trained in a fully supervised setting and frozen while we also use the pseudo labels to continuously updating the convolutional stream in our framework.

\textbf{Semi-supervised learning.} Effective supervised learning using deep neural networks usually requires annotating a large amount of data. However, creating such large datasets is costly and labor-intensive. A promising solution is SSL, which leverages unlabeled data to improve model performance. 
Existing SSL methods are designed from the aspects of pseudo labeling where model predictions are converted to hard labels (\eg, \cite{lee2013pseudo, rosenberg2005semi,yang2021deep}), and consistency regularization where the model is constrained to have consistent outputs under different perturbations \cite{bachman2014learning,rasmus2015semi, tarvainen2017mean,xie2020unsupervised, berthelot2019remixmatch}. FixMatch \cite{sohn2020fixmatch} combines these two classic semi-supervised learning strategies. It predicts hard pseudo labels under weak perturbations and guides the model to learn on unlabeled data with strong perturbations. Our work is built upon FixMatch to explore the potential of semi-supervised Vision Transformers. The noisy student \cite{Xie_2020_CVPR} extends the idea of self-training and distillation with larger student models and add noise to the student.  \cite{zhang2020pushing} applies transformers to automated speech recognition using semi-supervised learning. Their superior performance is obtained by large scale pre-training and iterative self-training using the noisy student training approach. 

As the advances of self-supervised learning approaches \cite{caron2021emerging,chen2020simple}, a new trend for semi-supervised learning becomes first utilizing the large scale unlabeled data for self-supervised pre-training and then use the labeled data for fine-tuning. Chen \etal \cite{chen2020big} show that a big ResNet pre-trained using SimCLRv2 can achieve competitive semi-supervised performance after fine-tuning.

\textbf{Joint modeling of CNNs and Transformers.}
CNNs and Transformers use two different ways to enforce geometric structure priors. A convolution operator is applied on patches of an image, which naturally results in a local geometric inductive bias. However, a Vision Transformer model utilizes the global self-attention to learn the relationships between global image elements \cite{dosovitskiy2020image}. 
From a complementary point of view, combining the advantages of CNNs in processing local visual structures and the advantages of transformer in processing global relationships is potentially a better approach for image modeling. 

One research direction is to imitate the CNN operations into a Vision Transformer or vice versa \cite{heo2021rethinking, yuan2021tokens, xiao2021early, wang2021pyramid}. For example, Pooling-based Vision Transformer (PiT) \cite{heo2021rethinking} applies pooling operations to shrink the feature maps and gradually increases the channel dimension at the same time, similar to the practice of CNN. PyramidViT~\cite{wang2021pyramid} and CvT~\cite{wu2021cvt} also adopt a similar hierarchical design. T2T-ViT \cite{yuan2021tokens} designs a progressive tokenization module to aggregate neighboring tokens. ~\cite{xiao2021early} replaces the ViT stem by a small number of stacked convolutions and observes it improves the stability of model training. They also keep the network deep and narrow, inspired by CNNs.

Probably the most relevant approaches are \cite{wang2018non, d2021convit, gulati2020conformer, peng2021conformer} that aim to find ways to combine convolution and transformer into a single model. For example, the non-local network \cite{wang2018non} adds self-attention layers to CNN backbones. SpeechConformer \cite{gulati2020conformer} attempts to use convolution to enhance the capabilities of the transformer, while ConVit \cite{d2021convit} introduces gated positional self-attention (GPSA) module which becomes equipped with a ``soft'' convolutional inductive bias. VisualConformer \cite{peng2021conformer} decouples CNN and Vision Transformer streams and design a module for feature communication across streams. However, these studies are all focused on supervised learning while we propose a generic framework for training semi-supervised vision transformers. Another major difference lies in that, even though we follow the same direction of fusing convolutions and transformers, our approach does not treat the combined architecture as an entirety, \eg, the pseudo labels have to be generated by the convolutional stream only.

\begin{figure*}[t]
\centering
\includegraphics[width=1.0\linewidth]{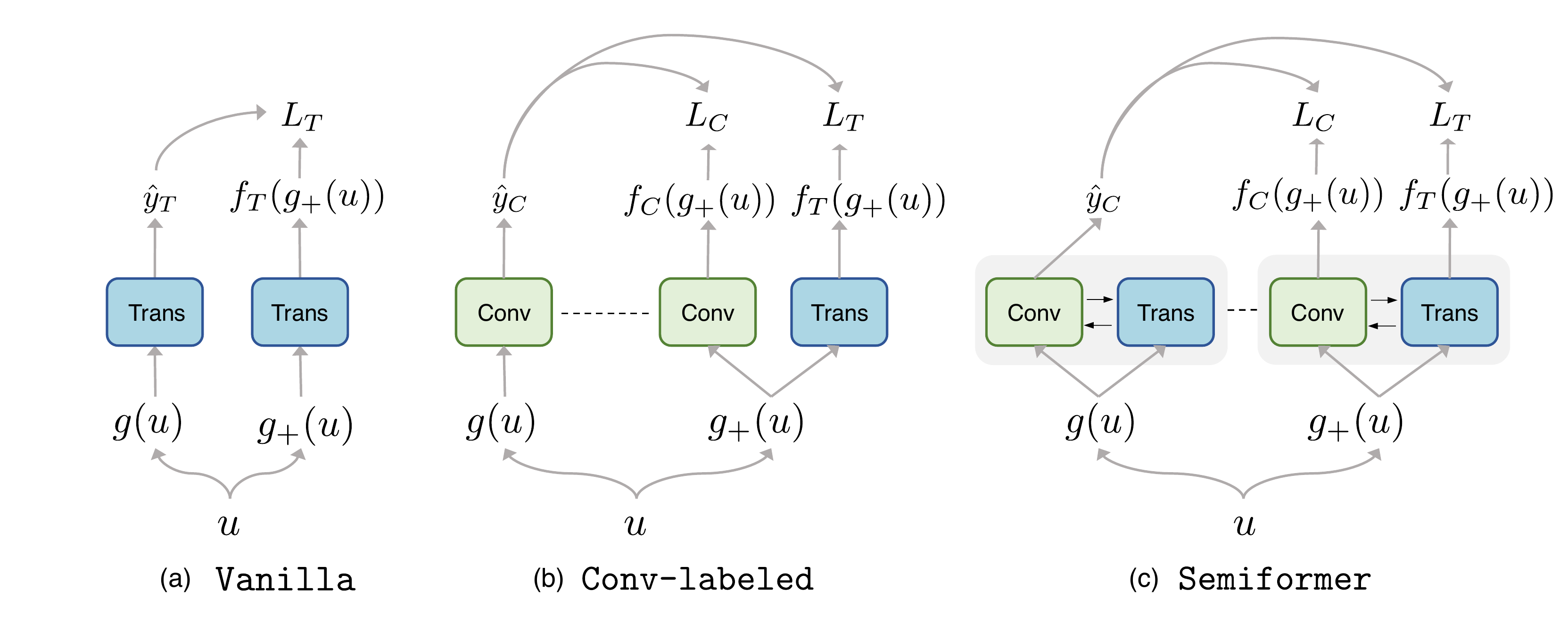}
  \caption{We explore a variety of ways to apply vision transformer into semi-supervised learning task. Dotted line refers to weights sharing. $u$ refers to the input image, $g$ and $g_{+}$ refer to weak and strong data augmentation. $\hat{y}_T$ and $\hat{y}_C$ refer to pseudo labels produced by transformer and convolutional streams. $f_T(\cdot)$ and $f_C(\cdot)$ represent model predictions of transformer and convolutional streams respectively.}
\label{fig:study}
\end{figure*}

\section{A Study with Vision Transformers for SSL} \label{sec:study}
We start by presenting two frameworks that use pseudo labels for SSL. Although the two attempts are surprisingly unsatisfactory, their results reveal two important lessons which eventually inspire us to develop our framework. Below we provide the details of the two studies and our learned lessons. 

\fakeparagraph{Unlabeled data improves Vision Transformers.} A natural approach to leverage unlabeled data is to do pseudo labeling through Vision Transformers. Our first hypothesis is that \emph{a Vision Transformer can be improved when the total number of input-output training pairs increases (though many of them are pseudo labels)}. We verify this with a {\vanilla} framework, which uses the same architecture (\eg, CNN or Transformer) and builds upon FixMatch~\cite{sohn2020fixmatch} for SSL. In particular, FixMatch uses two types of augmentations, a strong one and a weak one. The pseudo label of the unlabeled data is obtained by applying the model on weakly augmented images. And the model is trained using the strongly augmented inputs with the pseudo labels. 

\begin{table}[h]
\renewcommand\arraystretch{1.1}
\centering\small
\caption{Results and comparisons with two different SSL frameworks, and comparisons with the supervised baselines.}
\setlength{\tabcolsep}{10pt}
\begin{tabular}{cc cccc}
\toprule
\textbf{Architecture} & \textbf{Method} & \textbf{Top-1 Acc (\%)} \\
\midrule
\multirow{2}{*}{CNN} & \supervised only (10\%) & 60.2  \\
& \vanilla & \textbf{68.5}  \\
\midrule
\multirow{3}{*}{Transformer} & \supervised only (10\%) & 48.6 \\
& \vanilla & 59.0 \\
& \convl & 67.2 \\
\bottomrule
\end{tabular}
\label{table:preexp}
\end{table}
\setlength{\tabcolsep}{1.4pt}

Results in \cref{table:preexp} show that after adding the other 90\% images from the ImageNet as unlabeled training data, the transformer-based model can have an accuracy improvement by 10.4\%, which is greater than the accuracy improvement of CNN's 8.3\%. This validates our hypothesis, \ie, large-scale data helps the Vision Transformer to learn better even when many of them are pseudo labeled. However, despite the score increases, the performance of Vision Transformers in semi-supervised learning is still unsatisfactory, even inferior to the accuracy of fully supervised CNN training on only 10\% of the labeled data.

\fakeparagraph{Pseudo labels from CNNs are more accurate.}  We suspect that \emph{the weak performance of} \vanilla \emph{is due to the inaccurate pseudo labels generated by the transformer}. Vision Transformer contains less image-specific inductive bias, leading to poor performance on small-scale data and thus requires more data for representation learning.
In contrast, CNNs are shown to possess strong image-specific inductive bias due to its convolution and pooling design. A natural question is: \emph{what if we use a CNN to generate pseudo labels for Vision Transformer?} 

We introduce a new SSL framework, \convl,  which uses labels from CNNs for the SSL of CNN and transformers jointly, as illustrated in \cref{fig:study}(b). As is seen in \cref{table:preexp}, the \convl approach results in 67.2\% top-1 accuracy using the predictions from the ViT on ImageNet, improving the \vanilla approach by 8.2\%, which suggests that CNNs provide better pseudo labels.

\fakeparagraph{Conv-based pseudo labeling is not enough.} Although the ViT's performance is boosted by a CNN pseudo-label generator, the final performance of the ViT (67.2\%) is still worse than the CNN (68.5\%), observed from \cref{table:preexp}. This suggests that the knowledge from the CNN is not yet fully utilized through the simple pseudo labeling approach. 
One major problem here is that the two models are mostly decoupled except for the unilateral supervision given by the CNN. On the one hand, knowledge from the CNN is not directly injected into the transformer model. On the other hand, the CNN does not gain any information from the ViT.
This motivates us to consider jointly modeling both a convolution network and a transformer, which becomes the proposed \system framework.

\begin{figure}[t!]
\centering
\includegraphics[width=1.0\linewidth]{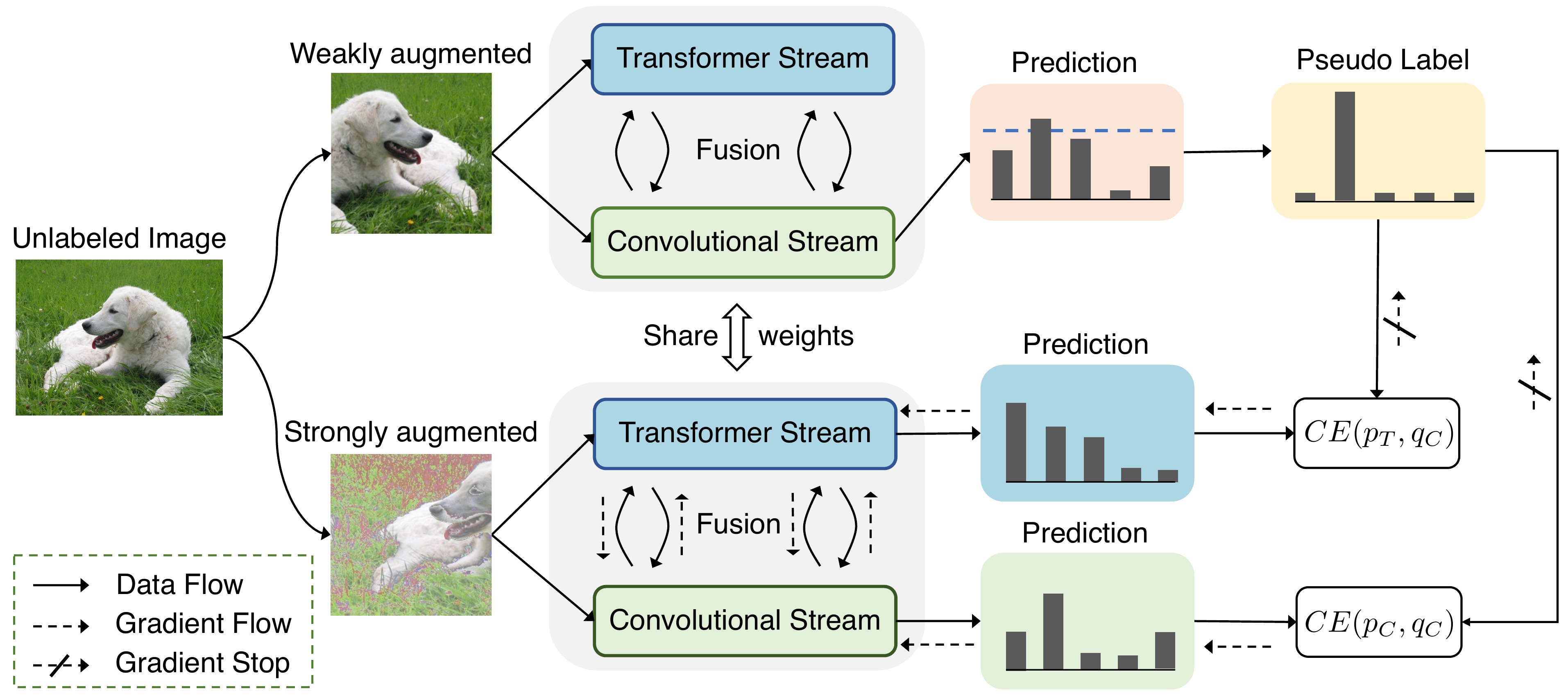}
  \caption{Diagram of the \system framework. For an unlabeled image, its weakly-augmented version (top) is fed into the model. The prediction of CNN is used for generating pseudo labels with a confidence threshold (dotted line). Then we compute the model's prediction for the strong augmented version of the same image (bottom). We expect both transformer and convolutional streams to match the pseudo label via cross-entropy losses. Streams complement each other with the feature-level modules.
  }
\label{fig:framework}
\end{figure}

\section{Our Approach: \system}

We introduce {\system} (illustrated in \cref{fig:study}(c) and \cref{fig:framework}), which jointly fuses a CNN architecture and a Transformer for semi-supervised learning.

\fakeparagraph{Notation.} We use $f(x; \theta)$ to represent the mapping function of our \system, given the input $x$ and the model parameter $\theta$. $f_T(x)$ and $f_C(x)$ are the vectorized output probability for each label from the transformer stream and the convolutional stream, respectively, and $\theta$ is omitted for simplicity.  Additionally, a weak data augmentation function $g(\cdot)$ and a strong data augmentation function $g_+(\cdot)$ are used in our approach. We assume the semi-supervised dataset contains $N_l$ labeled examples and $N_u$ unlabeled examples. We use index $i$ for labeled data, index $j$ for unlabeled and index $k$ for the label space. 

\fakeparagraph{Loss for labeled data.} Formally, the total loss for labeled data is
\begin{equation}
\label{eq:new_loss1}\small
{\mathcal L}_l = \sum_{i=1}^{N_l}\mathcal L_{xent}(y_i,f_T(g_+(x_i)))+\mathcal L_{xent}(y_i,f_C(g_+(x_i))) 
\end{equation}
where $x_i$ is the $i$-th labeled example and $y_i$ is its corresponding one-hot label vector. $\mathcal L_{xent}$ is the cross-entropy loss function, \ie, $\mathcal L_{xent}(p,q)=\sum_k p_k\log q_k$.

\fakeparagraph{Loss for unlabeled data.} Given an unlabeled image $u_j$, we first perform strong data augmentation $g_+(\cdot)$ and weak data augmentation $g(\cdot)$, according to FixMatch \cite{sohn2020fixmatch}, to obtain the two views of the same input image. However, only the prediction output of the convolutional stream $f_C(g(u_j))$ is used to generate the pseudo label.
The probability $p_j$ of an unlabeled input $u_j$ becomes
\begin{equation}
\label{eq:new_pred}
p_j = f_C(g(u_j))~.
\end{equation}
We define the pseudo labels as the class with maximum probability, \ie, $\hat p_j=\arg\max_k p_{jk}$. We use $\hat y_j$ to represent the one-hot vector corresponding to pseudo label $\hat p_j$. These pseudo labels will in turn be used to calculate the cross entropy loss to back-propagate both the convolutional and the transformer streams with the strongly augmented inputs $g_+(u_j)$.  A filtering by threshold $\max_k p_{jk}\ge\tau$, equivalent to $\langle\hat y_j,p_j\rangle\ge\tau$, is applied to remove pseudo labels without sufficient certainty. The remaining pseudo labels are used to guide the semi-supervised learning. 
The total loss for unlabeled data becomes
\begin{align}
\label{eq:new_lossu_T}
\small\mathcal L_u = \sum_{j=1}^{N_u}&\left(\mathcal L_{xent}(\hat y_j,f_T(g_+(u_j)))
+\mathcal L_{xent}(\hat y_j,f_C(g_+(u_j)))\right)\delta[\langle\hat y_j,p_j\rangle\ge\tau]~,
\end{align}
where $\delta[\cdot]$ is the delta function whose value is 1 when the condition is met and 0 otherwise. 

\fakeparagraph{Total loss.} The total training loss is the sum of both labeled and unlabeled losses such that
\begin{equation}
\label{eq:usloss2}
\mathcal L = \mathcal L_l+\lambda\mathcal L_{u}~,
\end{equation}
where $\lambda$ is a trade-off. A more detailed study of $\lambda$ can be found in \cref{sec:discussion}.

\fakeparagraph{Stream fusion.} Let $M_T$ be the Vision Transformer feature map in a certain layer with the shape ($d_T, h_T, w_T$) representing depth, height and width, respectively. Let $M_{T,i}$ be the $i$-th patch feature according to $M_T$ with the shape ($d_T, 1, 1$). So, $M_{T,i}$ corresponds to a specific area of the original image and we denote the CNN sub-feature map who also corresponds to the same area as $M_{C,i}$ with the shape ($d_C, h_C, w_C$). Motivated by~\cite{peng2021conformer}, we exchange information between patch features and its related CNN sub-feature map, described as \cref{eq:pool} and \cref{eq:upsample}:
\begin{align}
M_{T,i}&\mathrel{+}= {\texttt{layernorm}}({\texttt{pooling}}(\texttt{align}(M_{C,i}))),
\label{eq:pool}\\
M_{C,i}&\mathrel{+}= {\texttt{batchnorm}}({\texttt{upsample}}(\texttt{align}(M_{T,i}))),
\label{eq:upsample}
\end{align}
where the \texttt{align} operator refers to mapping features to the same dimensional space, \texttt{pooling} refers to downsampling, \texttt{upsample} refers to upsampling, \texttt{layernorm} refers to layer normalization and \texttt{BN} refers to batch normalization.
Specifically, a \texttt{Conv1x1} layer is used for embedding dimension alignment (the \texttt{align} operator). The average pooling and spatial interpolation methods are used for spatial dimension alignment, \ie, the \texttt{pooling} operator and the \texttt{upsample} operator, respectively.

To summarize, our framework consists of two parts, including carrying out a hard-way distillation manner by a convolutional stream to guide the transformer's learning from unlabeled data, and carrying out feature-level information interaction between the two streams so that the CNN's knowledge can be injected into the transformer and the convolutional stream can also be enhanced with a better global spatial information organization capability.

\fakeparagraph{Inference.} During training, we use the pseudo labels derived by the convolutional stream to train both the CNN and the Vision Transformer in a semi-supervised setting. For inference, we simply average combine predictions from the both streams as final scores, which is slightly better than using the transformer stream alone, as will be shown empirically.

\section{Experiments}

\subsection{Experimental setup}

\fakeparagraph{Datasets and evaluation metrics.}  To evaluate the effectiveness of \system, we mainly conduct experiments on \inet \cite{deng2009imagenet}, which contains 1,000 classes and 1.3M images.  In addition, we provide experimental results on \places. Unlike \inet that contains generic categories, \places is a place-focused dataset, which contains 2.5M images annotated into 205 classes. We use top-1 accuracy as our evaluation metric.  Through all experiments, following~\cite{sohn2020fixmatch}, we mainly select 10\% labeled samples and leave the other 90\% samples as unlabeled data, unless specified otherwise. 

\fakeparagraph{Models.} 
The \system framework emphasizes how to complement the characteristics of the CNN and the ViT to achieve improved results. 
For the convolutional stream, we use a ResNet-like model and a personalized ConvMixer\cite{trockman2022patches}, while within transformer stream, we experiment with both a slightly modified ViT-S\cite{touvron2021training} and the PiT-S\cite{heo2021rethinking} as backbone networks.

\fakeparagraph{Implementation details.} The initial learning rate is set to $10^{-3}$ and is decayed towards $10^{-5}$ following the cosine decay scheduler. We use 5 epochs to warm-up our models and another 25 epochs to train models on the labeled data before starting the semi-supervised learning process. In the training of ViT-ConvMixer model, the batch size of each GPU is 84, while in the training of ViT-Conv and PiT-Conv model, the batch size is 108 per GPU.  We train models with 600 epochs using 32 NVIDIA V100 GPUs to produce our best top-1 accuracy by setting the number ratio of labeled and unlabeled images in each batch as 1:7. In order to avoid gains brought by data augmentation, we do not apply mixup, cutmix and repeat augmentation in our SSL process. We choose random augmentation, random erasing and color jitter as the strong data augmentation, and use random flipping and random cropping as the weak data augmentation. The value of $\lambda$ which is the balance factor between loss terms is set as 4.0. In the semi-supervised learning with 5\% \inet labeled samples, we reduce the number ratio of labeled and unlabeled images per batch to 1:9. All the experiments share the same G.T. data split.

For ablation studies and discussion, we train 300 epochs to speed up the experiments and we set the number ratio of labeled and unlabeled images in each batch as 1:5 and use the label smoothing trick on ground-truth labels.

\subsection{Main Results}

\fakeparagraph{Comparisons with state-of-the-art.} We first compare with state-of-the-art semi-supervised methods, such as UDA~\cite{xie2020unsupervised}, FixMatch~\cite{sohn2020fixmatch}, S4L~\cite{zhai2019s4l}, MPL~\cite{pham2021meta} and CowMix~\cite{french2020milking}, as well as recent self-supevised methods. Experimental results in \cref{table:sota} show that our approach achieves better results by clear margins compared with alternative methods. For example, \system is better than S4L~\cite{zhai2019s4l} and CowMix~\cite{french2020milking} by 2.3\% and 1.6\% with only 11\% and 67\% of parameters of their models, respectively. In addition, while we follow the design the principle of FixMatch to generate pseudo labels, the knowledge sharing mechanism in \system brings about 4\% performance gain compared to FixMatch.  Although MPL has a smaller model size, training MPL is computationally expensive as it requires meta updates. In addition, MPL uses complicated data augmentations, \ie, AutoAugment, while we only use basic augmentations. Similarly, CowMix~\cite{french2020milking} introduces a new data augmentation strategy for SSL. We would like to point that \system is a generic SSL framework that explores pseudo labels and knowledge in CNNs to promote the results of transformers.  We believe it is in tandem with more advanced pseudo label generation strategies like MPL~\cite{pham2021meta} and more complex augmentation methods~\cite{french2020milking}. In addition to SSL methods, we also compare with self-supervised learning results such as~\cite{henaff2020data,chen2020big,grill2020bootstrap}, which firstly learn representations with self-supervised methods and then perform finetuning on limited data. We see that \system also performs favorably compared to this line of methods.

\setlength{\tabcolsep}{10pt}
\begin{table}[t]
\renewcommand\arraystretch{1.0}
\centering
\caption{The results of \system and comparisons with state-of-the-art methods. \system achieves 75.5\% top-1 accuracy and outperforms all Convolutional neural network based methods, while still keeping a reasonable parameter size. Here, the params does not include the final classifier.}
\label{table:sota}
\resizebox{.93\textwidth}{!}{
 \setlength{\tabcolsep}{0pt} 
   \begin{tabular*}{\linewidth}{@{\extracolsep{\fill}\quad}lccc}
\toprule
 \multicolumn{1}{c}{\textbf{Method}}  & \textbf{Architecture} & \textbf{Params} & \textbf{Top-1 Acc(\%)}\\
\midrule
  \multirow{2}{*}{\supervised (10\%)} & ViT-S  & 23M & 48.6 \\
   & Conv & 13M & 60.2 \\
\midrule
\multicolumn{1}{l}{\small{\textit{Self-supervised pretraining}}} \\
  CPC~\cite{henaff2020data} & ResNet-161 & 305M & 71.5 \\
 SimCLR~\cite{chen2020simple,chen2020big}& ResNet-50 & 24M & 65.6 \\
 SimCLR~\cite{chen2020simple,chen2020big}&ResNet-50 (2×) & 94M & 71.7\\
 BYOL~\cite{grill2020bootstrap} & ResNet-50 & 24M & 68.8 \\
 BYOL~\cite{grill2020bootstrap} & ResNet-50 (2×) & 94M & 73.5 \\
 DINO~\cite{caron2021emerging} & ViT-S & 21M & 72.2 \\
\midrule 
\multicolumn{3}{l}{\small{\textit{Semi-supervised methods}}} \\
 UDA\cite{xie2020unsupervised} & ResNet-50 & 24M & 68.8  \\
 FixMatch\cite{sohn2020fixmatch} & ResNet-50 & 24M & 71.5  \\
 S4L  \cite{zhai2019s4l} & ResNet-50 (4×) & 375M &  73.2 \\ 
 MPL \cite{pham2021meta} &  ResNet-50 & 24M & 73.9  \\
 CowMix \cite{french2020milking} & ResNet-152 & 60M & 73.9 \\
\midrule
  \multirow{ 1}{*}{\system} & ViT-S + Conv  & 40M & \textbf{75.5} \\
\bottomrule
\end{tabular*}
}
\end{table}

\setlength{\tabcolsep}{1.4pt}

\setlength{\tabcolsep}{6pt}
\begin{table}[t]
\renewcommand\arraystretch{1.1}
\centering
\caption{Ablation Study: The effectiveness of \system with various backbones and comparisons with alternative methods (\ie, vanilla and conv-labeled). All models are trained with 300 epochs and without pseudo label smoothing. For \convl and \system, A/B in the last column: A indicates scores from the transformer stream only and B indicates averaged predictions from CNNs and transformers. }

\resizebox{0.93\textwidth}{!}{
   \setlength{\tabcolsep}{0pt} 
\begin{tabular*}{\textwidth}{@{\extracolsep{\fill}\quad}lccc}
\toprule
\multicolumn{1}{c}{\textbf{Method}} & \textbf{Backbone} & \textbf{Pseudo labels} & \textbf{Top-1 Acc(\%)} \\
\midrule 
 \multirow{ 2}{*}{\supervised (10\%)}& ViT-S & - & 48.6\\
 & PiT-S & - &  50.0 \\
\midrule 
 \multirow{ 4}{*}{\vanilla}& Conv &  Conv & 68.5  \\
 & ConvMixer & ConvMixer & 69.3 \\
 &ViT-S & ViT-S & 59.0\\
 & PiT-S & PiT-S &  63.0 \\
\midrule
\multirow{ 3}{*}{\convl} & ViT-S + Conv & Conv &  67.2 / 70.2 \\
&  PiT-S + Conv & Conv  & 67.8 / 70.5 \\
 & ViT-S + ConvMixer & ConvMixer &  66.7 / 70.2 \\
\midrule
\multirow{ 3}{*}{\system} & ViT-S + Conv & Conv &   72.4 / 73.5 \\
&  PiT-S + Conv & Conv  &  70.8 / 71.6  \\
 &   ViT-S + ConvMixer  & ConvMixer &   72.9 / \textbf{73.8} \\
\bottomrule
\end{tabular*}
}
\label{table:analysis}
\end{table}
\setlength{\tabcolsep}{1.4pt}

\fakeparagraph{Effectiveness of \system with different backbones.} We evaluate the performance of \system instantiated with different CNN and transformer backbones using 10\% of labeled samples. We compare with the supervised training baseline (\supervised), the \vanilla method where the pseudo label generator share the same backbone used for SSL, and \convl that trains transformers with labels produced by CNNs.  The results are summarized in \cref{table:analysis}. As the \vanilla results shown in the second block of \cref{table:analysis}, CNNs obviously achieve higher image classification accuracy than Vision Transformers under the SSL setting, verifying that labels from CNNs are more accurate. ConvMixer~\cite{trockman2022patches} achieves the best results among all \vanilla models, offering a top-1 accuracy of 69.3\%. This possibly results from the fact ConvMixer integrates the architectural advantages of both transformers and CNNs. Results in the third block of \cref{table:analysis} show that using CNNs instead of Vision Transformers to generate pseudo labels significantly improves the performance of the Vision Transformer, allowing PiT-S and ViT-S to reach an accuracy of 67.2\% and 67.8\% respectively, with the same CNN architecture. The improved accuracy is close to that of the \vanilla semi-supervised CNN, suggesting the quality of the pseudo labels makes a difference to the semi-supervised learning process of Vision Transformers. 

Results in the last block of \cref{table:analysis} show that \system significantly improves the performance of Vision Transformers. This highlights the effectiveness of \system in exploring the interactions of CNNs and transformers. Taking the combination of ViT-S and Conv as an example, after applying the feature-level interaction to accomplish the dual information exchange, the accuracy of ViT-S is improved by 5.2\% from 67.2\% to 72.4\%, revealing the efficacy of our \system framework. We also observe that \system is a versatile framework compatible with modern CNN and transformer architectures.  In addition, by further combining the predictions from both the convolutional and transformer streams, we observe consistent performance gains under all settings for \convl and \system.

\fakeparagraph{\bf The ratio of labeled samples.} 
we further experiment with 5\% and 20\% of labeled samples for SSL and compare with alternative methods. Except that we decrease the number of labeled and unlabeled images in each batch from 1:5 to 1:9  for 5\% labeled samples, all the experimental settings are kept the same as those of using 10\% labeled samples. \cref{table:ratio} presents the results. Vision transformer performs poorly when only 5\% labels are available, with an accuracy of only 28.6\%, which is 15.6\% lower than the Conv accuracy of 44.2\%. With the increase of the number of labeled samples, the performance gain of ViTs is more significant than that of CNNs. For example, with ViTs, the accuracy increases by 20\% and 14.3\% respectively, when the number labeled samples grows from 5\% to 10\% and from 10\% to 20\%, respectively, suggesting that the training of Vision Transformers is more sensitive to the number of labels. In addition, we see that pseudo labels from CNNs are more accurate and help ViT learn better.

\setlength{\tabcolsep}{6pt}
\begin{table}[t]
\renewcommand\arraystretch{1.2}
\centering
\caption{SSL with different ratios of labeled samples on \inet.}
\resizebox{.95\textwidth}{!}{
\begin{tabular}{c|c|cc|cc|cc}
\toprule
  \multirow{2}{*}{Dataset} & \multirow{2}{*}{Ratio}& \multicolumn{2}{c|}{ViT-S} & \multicolumn{2}{c|}{Conv} & \multicolumn{2}{c}{ViT-S + Conv} \\
  && \supervised & \vanilla &  \supervised & \vanilla &\convl & \system \\
\midrule
  \multirow{3}{*}{\inet}&5 \% & 28.6 & 45.7 \Rise{17.1} & 44.2 & 61.3 \Rise{17.1} & 62.0 & 66.3 \Rise{4.3}  \\
  &10 \% & 48.6 & 59.0 \Rise{10.4} & 60.2 & 68.5 \Rise{ 8.3} & 70.2 & 73.5 \Rise{3.3} \\
  &20 \% & 52.9 & 69.8 \Rise{16.9} & 63.5 & 73.6 \Rise{10.1} & 74.8 & 78.1 \Rise{3.3}  \\
\bottomrule
\end{tabular}
}
\label{table:ratio}
\end{table}
\setlength{\tabcolsep}{1.4pt}

\fakeparagraph{\bf Extension to Places205.} We also conduct experiments on \places to further evaluate the effectiveness of \system. 
As \places is roughly 2 times larger than \inet, we use 5\% of labeled samples to assure the semi-supervised experiments on 5\% \places and 10\% \inet have approximately the same number of labeled samples.  We see from \cref{table:places} that \system consistently produces the best results. For example, \system is 1.3\% and 6.9\% better than \convl and \vanilla-ViT-S, respectively. Similar trends can be observed by comparing across \cref{table:ratio} and  \cref{table:places}, which further confirms the efficacy of \system.

\setlength{\tabcolsep}{6pt}
\begin{table}[t]
\renewcommand\arraystretch{1.2}
\centering
\caption{Top-1 Accuracy of \system on 5\% labeled subset of Places205.}
\resizebox{.95\textwidth}{!}{
\begin{tabular}{c|c|cc|cc|cc}
\toprule
  \multirow{2}{*}{Dataset} & \multirow{2}{*}{Ratio}& \multicolumn{2}{c|}{ViT-S} & \multicolumn{2}{c|}{Conv} & \multicolumn{2}{c}{ViT-S + Conv} \\
  && \supervised & \vanilla &  \supervised & \vanilla &\convl & \system \\
\midrule
  \multirow{1}{*}{\places}&5 \% & 36.0 & 46.9 \Rise{3.9} & 44.3 & 51.6 \Rise{7.3} & 52.5 & 53.8 \Rise{1.3}  \\
\bottomrule
\end{tabular}
}
\label{table:places}
\end{table}
\setlength{\tabcolsep}{1.4pt}

\begin{figure}[t]
\centering
  \includegraphics[width=1.0\linewidth]{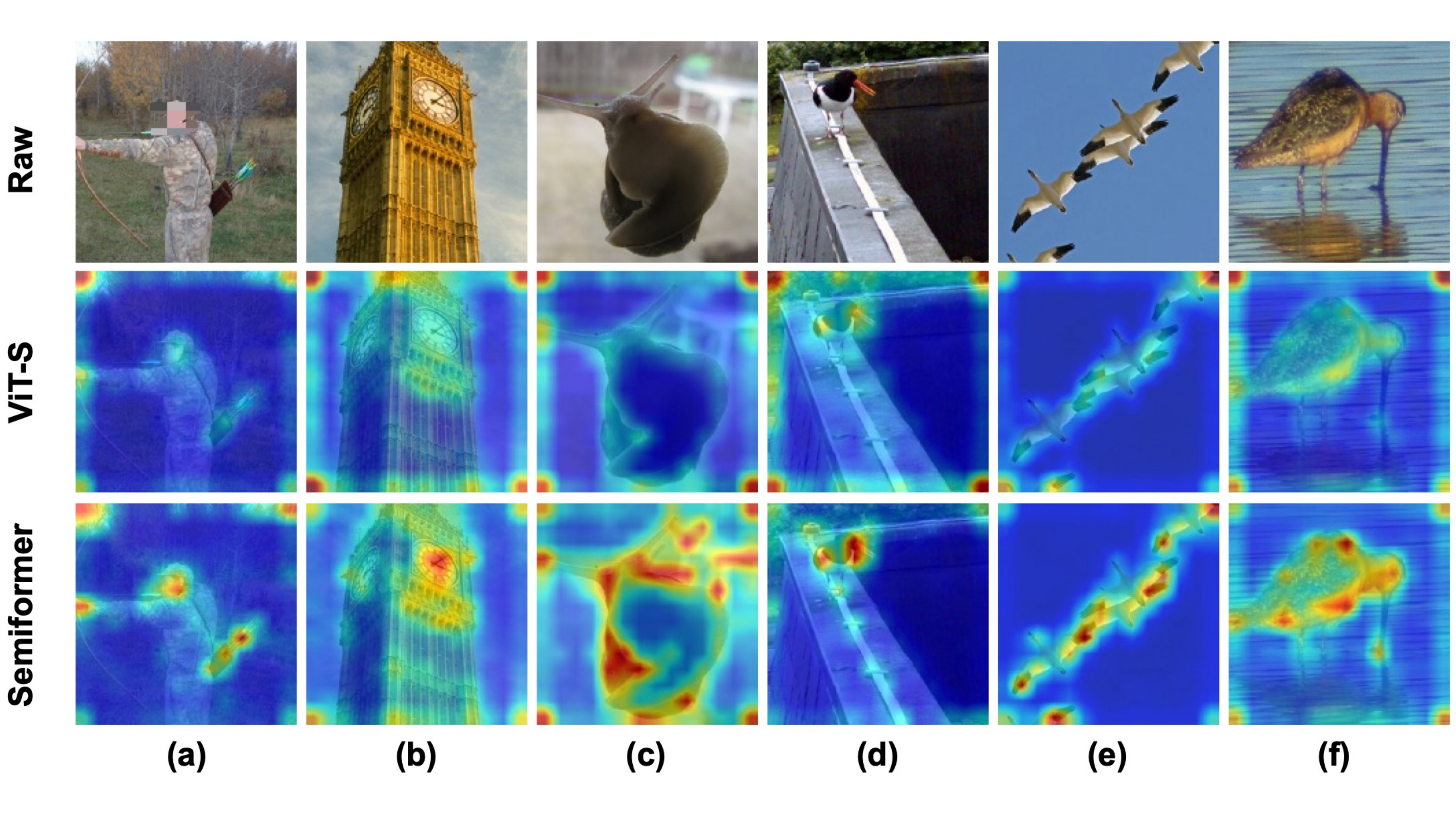}
  \caption{Attention map of ViTs and \system using samples from ImageNet. Compared to ViTs where the attention scores are scattered, \system focuses more on critical objects.}
\label{fig:attn_visualization}
\end{figure}

\subsection{Qualitative Results} We visualize in \cref{fig:attn_visualization} the attention maps of ViT and \system.
Thanks to the guidance of pseudo label generator CNN and its supplementary help of injecting the local information extraction ability, \system can retain more local information of images and can correctly focus on the key local positions of the images. For example, when analyzing the \cref{fig:attn_visualization}(a) which corresponds to the class of \emph{bow}, \system is particularly more focused on the man's hand holding the bow, the man's head and the quiver carried by the person, and those attended areas are critical for identifying the bow category. In addition, \system covers essential objects precisely. In \cref{fig:attn_visualization}(f), we can see the attention map of \system not only covers the animal completely, but also covers the contours more tightly. And for images with many small objects, for instance, \cref{fig:attn_visualization}(e), \system shows stronger ability to concentrate on key local areas and coverage the essential areas.

\subsection{Discussion}\label{sec:discussion}
\fakeparagraph{\bf What model should be used to produce pseudo labels?} Although models in our \system framework interact with each other, the CNN model still outperforms the vision transformer especially in the early training stage, making it important to retain the CNN hard-way distillation mode. To verify this, we replace the teacher stream which is responsible for generating pseudo labels. We use the following three strategies to produce pseudo labels: CNNs only, transformers only, and averaged predictions from CNNs and transformers. As shown in \cref{table:pseudo-type}, using the CNNs as the teacher network brings the highest accuracy, \ie 73.5\%, while using the transformer stream to generate pseudo labels performs worst (\ie, 67.4\%) .  As the quality of pseudo labels generated by vision transformers are limited, we do not get better results by simply averaging CNN and Vision Transformer outputs as pseudo labels under the same setting. This further confirms the effectiveness of our pseudo labeling strategy.

\fakeparagraph{\bf Does \system performs well because of larger models?} To clear up the confusion on the relationship between the number of parameters and  accuracy, we ablate on the model architecture of \system using different backbones. We experiment with different versions of ResNet~\cite{he2016deep} including ResNet-50 (R50), ResNet-101 (R101), ResNet-152 (R152). Results are presented in \cref{table:modelparam}. We observe that by adding more layers to ResNet, the top-1 accuracy of semi-supervised learning does gradually increase. However, it is still lower than that of \system. Even though the ResNet152 model contains 18M more parameters than \system, its accuracy is still 1.7\% worse than that of \system, which proves the performance gain of \system does not come from model sizes. We further instantiate the two streams of \system with the same backbone, \ie C+C and V+V respectively, and modify the stream connection correspondingly. Note that this is different from \vanilla as the two streams exchange information.  \cref{table:modelparam} reveals that these combinations are significantly worse than \system. For example, \system outperforms V+V by 13.9\% with 7M fewer parameters, which again shows the effectiveness of \system is not due to extra parameters.

\begin{figure}[t!]
\centering

\renewcommand\arraystretch{1.1}
\begin{minipage}[t]{.27\textwidth}
  \centering
    \setlength{\tabcolsep}{3pt}
    
    \renewcommand\arraystretch{1.0}
    \centering
         \tabcaption{Results by different pseudo labels.}
    \resizebox{1.0\textwidth}{!}{
      \begin{tabular}{cc}
        \toprule
        PL Type & Acc@1(\%)  \\
        \midrule
        CNN  & \textbf{73.5}  \\
        Trans  & 67.4  \\
        Fusion &71.1   \\
        \bottomrule
      \end{tabular}
      }
    \label{table:pseudo-type}
    
    \setlength{\tabcolsep}{1.4pt}
\end{minipage}
\quad
\begin{minipage}[t]{.65\textwidth}
    \renewcommand\arraystretch{1.2}
    \centering
    \tabcaption{Model size analysis. V and C refer to ViT-S and CNN, respectively. R represnets ResNet.}
    \resizebox{1.0\textwidth}{!}{
    \begin{tabular}{c|ccc|ccccc}
    \toprule
      Architecture & R50 & R101 & R152 & C & V & C+C & V+V & V+C\\
    \midrule
      Params & 24M & 43M & 58M & 13M & 23M & 35M & 47M & 40M \\
      Top-1 Acc(\%)  & 68.3 & 70.8 & 71.8 & 68.5 & 59.0 & 66.9 & 59.6 & \textbf{73.5} \\
    \bottomrule
    \end{tabular}
    }
    \label{table:modelparam}
\end{minipage}

\end{figure}

\fakeparagraph{\bf The impact of hyper-parameters.}
The default  set of hyperparameters are: label and unlabeled data ratio is 1:5, confidence threshold is 0.7 and $\lambda$ is set as 4. Based on the default setting, we control other variables unchanged and observe how the accuracy rate changes after independently changing the following three factors: different confidence threshold (0.65, 0.7, 0.75, 0.8); different $\lambda$ value (1, 2, 3, 4); different proportion of the number of labeled and unlabeled data (1:3, 1:5, 1:7). \system offers the best results with 0.7 confidence threshold, 1:7 labeled-unlabeled ratio, and $\lambda=4$.

\section{Conclusion}
We presented \system, the first framework to train Vision Transformers for semi-supervised learning. We found directly training a \vanilla transformer on semi-supervised data is ineffective. The proposed framework combines a CNN and a Vision Transformer using a cross fusion approach. The optimal semi-supervised learning performance is achieved by using only the convolutional stream to generate the pseudo labels. The final fused framework achieves 75.5\% top-1 accuracy on ImageNet and outperforms the state-of-the-art in semi-supervised image classification. 

\paragraph{\bf{Acknowledgement}} Y.-G. Jiang was
sponsored in part by ``Shuguang Program'' supported by Shanghai Education Development Foundation and Shanghai Municipal
Education Commission (No. 20SG01). Z. Wu was supported by NSFC under Grant No. 62102092.

\clearpage
%
%
\bibliographystyle{splncs04}
\bibliography{egbib}
\end{document}